\documentclass[lettersize,journal]{IEEEtran}

\usepackage{graphicx}
\usepackage{times}
\usepackage{epsfig}
\usepackage{epstopdf}
\usepackage{amsmath}
\usepackage{amssymb}
\usepackage{subfigure}
\usepackage{caption}
\usepackage{tabularx}
\usepackage[table]{xcolor}
 \usepackage[normalem]{ulem}
\usepackage{algorithm}
\usepackage[noend]{algpseudocode}

\usepackage[section]{placeins}

\usepackage{multirow}
\usepackage{amssymb}
\usepackage{booktabs}
\usepackage{soul}
\usepackage{lipsum}
\usepackage{enumitem}
\setlist[itemize]{leftmargin=*}
\usepackage{makecell}
\usepackage{graphicx}
\usepackage{animate}
\usepackage{setspace}
\usepackage{makecell}
\usepackage{array}
\definecolor{LightCyan}{rgb}{0.88,0.95,1}
\definecolor{mygreen}{HTML}{007F7F}
\definecolor{myred}{HTML}{CC0000}
\definecolor{myblue}{RGB}{0, 71, 171}
\definecolor{gold}{rgb}{0.75, 0.5, 0.25}

\makeatletter
\def\BState{\State\hskip-\ALG@thistlm}
\makeatother
\definecolor{myGreen}{HTML}{33FF00}
\definecolor{myRed}{HTML}{FF3030}
\definecolor{myGrey}{HTML}{AA5555}
\definecolor{myWhite}{HTML}{FFFFFF}
\definecolor{maroon}{cmyk}{0,0.87,0.68,0.32}
\definecolor{petr}{HTML}{5555FF}
\definecolor{josef}{HTML}{FF3030}

\begin{document}

\title{PEMF-VTO: Point-Enhanced Video Virtual Try-on via Mask-free Paradigm}

\author{Tianyu Chang$^{\ddag,\dag}$, Xiaohao Chen$^{\ddag}$, Zhichao Wei, Xuanpu Zhang, Qingguo Chen, Weihua Luo, Peipei Song, and Xun Yang$^*$,~\IEEEmembership{Member,~IEEE}
\thanks{$^*$Corresponding author: Xun Yang.}
\thanks{$^{\ddag}$These authors contributed equally to this work.}
\thanks{$^{\dag}$This author was an intern at Alibaba International Digital Commerce Group during the research.}
\thanks{Tianyu Chang, Peipei Song and Xun Yang are with the School of Information Science and Technology, University of Science and Technology of China (USTC), Hefei 230026, China (e-mail: cty8998@mail.ustc.edu.cn; beta.songpp@gmail.com; xyang21@ustc.edu.cn).}
\thanks{Xuanpu Zhang is with the School of Electrical and Information Engineering, Tianjin University, Tianjin 300072, China.}
\thanks{Xiaohao Chen, Zhichao Wei, Qingguo Chen, and Weihua Luo are with Alibaba International Digital Commerce Group, Hangzhou 311121, China.}
}

\markboth{IEEE TRANSACTIONS ON Consumer Electronics, 2025}%
{Chang \MakeLowercase{\textit{et al.}}: Point-Enhanced Video Virtual Try-on via Mask-free Paradigm}

\maketitle

\begin{abstract}
    Video Virtual Try-on aims to seamlessly transfer a reference garment onto a target person in a video while preserving both visual fidelity and temporal coherence. 
    Existing methods typically rely on inpainting masks to define the try-on area, enabling accurate garment transfer for simple scenes (e.g., in-shop videos). 
    However, these mask-based approaches struggle with complex real-world scenarios, as overly large and inconsistent masks often destroy spatial-temporal information, leading to distorted results. 
    Mask-free methods alleviate this issue but face challenges in accurately determining the try-on area, especially for videos with dynamic body movements. 
    To address these limitations, we propose PEMF-VTO, a novel Point-Enhanced Mask-Free Video Virtual Try-On framework that leverages sparse point alignments to explicitly guide garment transfer. 
    Our key innovation is the introduction of point-enhanced guidance, which provides flexible and reliable control over both spatial-level garment transfer and temporal-level video coherence. 
    Specifically, we design a Point-Enhanced Transformer (PET) with two core components: Point-Enhanced Spatial Attention (PSA), which uses frame-cloth point alignments to precisely guide garment transfer, and Point-Enhanced Temporal Attention (PTA), which leverages frame-frame point correspondences to enhance temporal coherence and ensure smooth transitions across frames. 
    Extensive experiments demonstrate that our PEMF-VTO outperforms state-of-the-art methods, generating more natural, coherent, and visually appealing try-on videos, particularly for challenging in-the-wild scenarios.  
\end{abstract}


\section{Introduction}
\label{sec:intro}

\IEEEPARstart{V}ideo Virtual Try-On, which aims to transfer the provided garment into a specific area in the source person video 
while maintaining the inter-frame coherence, 
has garnered significant attention  
especially for the E-Commerce and fashion design fields.  
This technology greatly reduces related costs and brings more convenient shopping and work experience to consumers and fashion designers.

\begin{figure*}[h]
    \centering
    \includegraphics[width=0.85\linewidth]{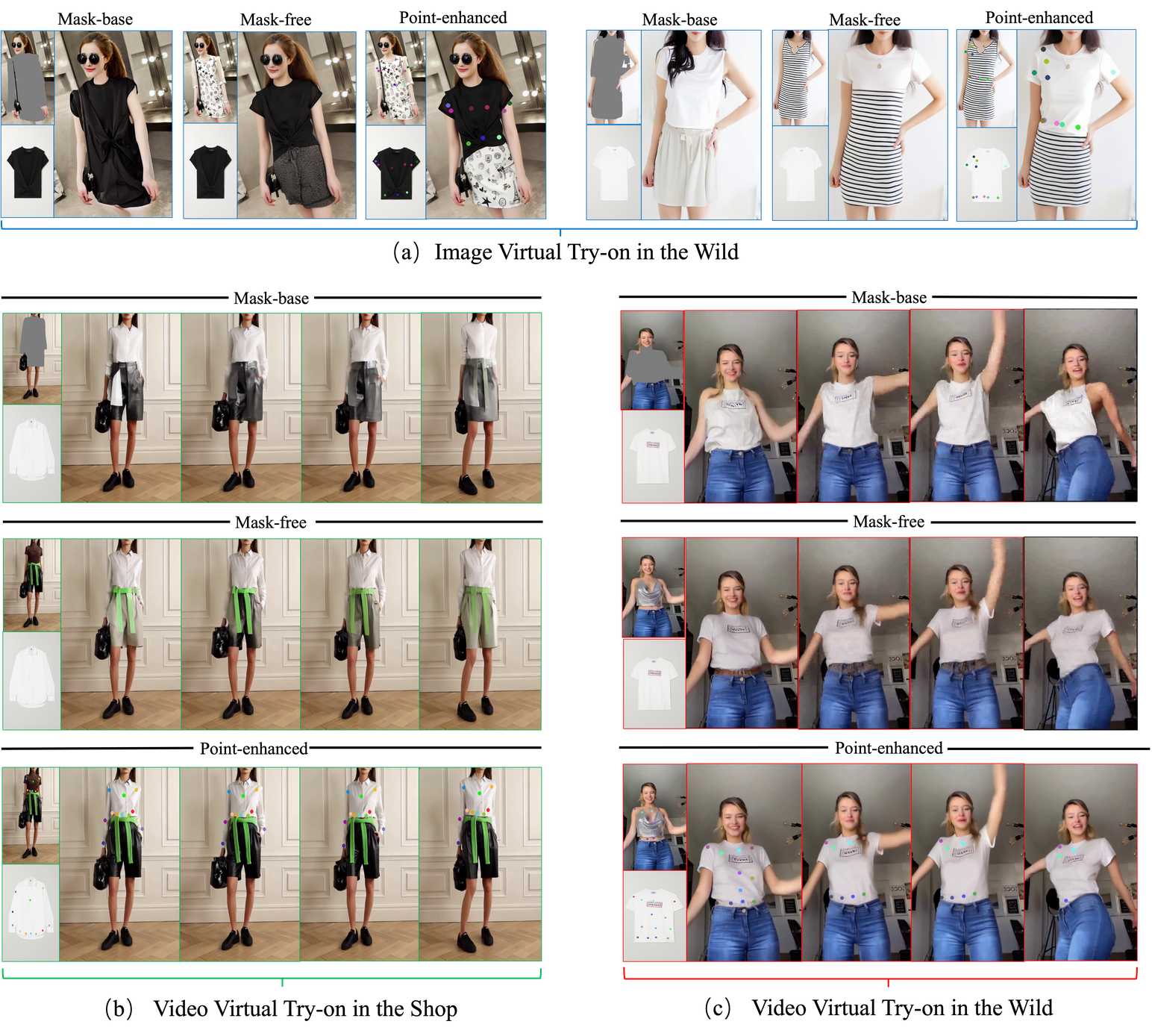}
    \caption{Comparison of the mask-based, mask-free and point-enhanced mask-free virtual try-on models. 
            Rows from top to bottom denote the (a) image try-on results in the wild, (b) video try-on results in the shop and (c) video try-on results in the wild. 
            The testing data are respectively from open-sourced StreetVTON~\cite{cui2023street-tryon}, ViViD~\cite{fang2024vivid} and TikTok~\cite{jafarian2021learning} datasets.}
    \label{fig1}
\end{figure*}

Recently, based on powerful diffusion models~\cite{ho2020denoising,nichol2021improved,song2020denoising,zhang2023adding} and existing image virtual try-on training paradigm ~\cite{choi2024improving,kim2024stableviton,ning2024picture,yang2024texture,zeng2024cat,zhu2024m}, 
many video virtual try-on methods~\cite{he2024wildvidfit,fang2024vivid,xu2024tunnel,wang2024gpd,zheng2024vitondit} have been proposed to take advantage of the inpainting mask to ensure the try-on area
and the temporal attention module to keep the video coherence, obtaining natural and impressive try-on results on simple in-shop videos.
However, in realistic scenarios, person videos often exhibit complex body movements and significant scene motion (\textit{e.g.}, street dance videos).
The pre-acquired agnostic mask of such challenging videos will lead to 1) the loss of spatial information on human postures and 2) temporal inconsistency in try-on areas between adjacent frames,
causing mask-based methods to generate distorted and incoherent try-on video, as Fig.~\ref{fig1} shown. 
Therefore, to learn a more general video virtual try-on model, 
it is necessary to design a more reasonable framework to eliminate the inherent deficiencies of mask-based try-on methods.

In the image virtual try-on area, several attempts have been made to alleviate the negative impact of the inpainting mask by either 
1) correcting the initial inaccurate agnostic mask~\cite{yang2024texture,zhang2024better} or 2) constructing the large-scale paired training data to learn a mask-free try-on model ~\cite{ge2021parser,he2022style,zhang2024boow,niu2024anydesign}.
However, the correlation of the agnostic mask does not eliminate the dependencies of the mask guidance.
In addition, as Fig.~\ref{fig1} shows, directly transferring the mask-free paradigm to a video virtual try-on task will bring confusing inconsistencies between reference garments and generated video frames.
The core reason is that person video data contain more complex actions or movements at the temporal level, compared to image data.
Based on the above analysis, \textit{it is necessary to design an innovative video virtual try-on paradigm that can avoid the deficiencies of agnostic masks while providing explicit and flexible guidance on the specific garment try-on area.}

In this work, we propose a novel \textbf{P}oint-\textbf{E}nhanced \textbf{M}ask-\textbf{F}ree \textbf{V}ideo virtual \textbf{T}ry-\textbf{O}n (PEMF-VTO) framework,
which exploits the sparse frame-cloth and frame-frame point alignments to explicitly guide the desirable try-on area while enhancing the coherence of the generated try-on video. 
Specifically, we first leverage the pre-trained mask-based try-on method ViViD~\cite{fang2024vivid} to construct large-scale paired data, thus learning a mask-free virtual try-on model that eliminates the negative impact of the unreliable agnostic mask. 
Inadequately, the pure mask-free model can not determine the consistent and correct try-on area at the temporal level, especially for realistic videos with diverse body movements.
To this end, we introduce sparse frame-cloth and frame-frame correspondences (matching point pairs) to enhance the consistency and coherence of garment transfer to source video.
Concretely, a \textbf{P}oint-enhanced \textbf{S}patial \textbf{A}ttention (PSA) is designed to strengthen the garment transfer to a specific area.
Besides, a \textbf{P}oint-enhanced \textbf{T}emporal \textbf{A}ttention (PTA) is also proposed to increase the coherence of the generated video.
In this way, our PEMF-VTO can simultaneously meet the three important and challenging requirements of the video virtual try-on task:
1) the accurate transfer of reference garment 2) the preservation of the non-try-on area and 3) the continuity of generated video frames.
Extensive qualitative and qualitative experiments clearly show the effectiveness of our method.
    
Our contributions can be briefly summarized as follows:
\begin{itemize}
 \item We investigate the deficiencies of current learning paradigms in virtual try-on methods and propose a more flexible and generalizable point-enhanced mask-free paradigm compared to prior approaches.
    \item We design the \textbf{P}oint-enhanced \textbf{S}patial \textbf{A}ttention (PSA) module and the \textbf{P}oint-enhanced \textbf{T}emporal \textbf{A}ttention (PTA) module to enhance the garment transfer ability and temporal coherence of the generated try-on video.
    \item Extensive experiments illustrate that our method achieves higher-quality and more coherent results for video virtual try-on, especially in challenging in-the-wild scenarios.
\end{itemize}

\section{Related work} 
\label{sec:related}
\noindent{\textbf{Image Virtual Try-On.}} 
Given a source person image and a reference garment image, image virtual try-on aims to synthesize an identity and background preserved and cloth changed image.
Previous methods~\cite{ge2021parser,zheng2021collocation,dong2022tryoncm2,he2022style,han2017viton,wang2018toward,ge2021parser,choi2021viton,dong2022dressing,han2019clothflow,yang2022full,zhang2021pise,10564147,10305193,10065574,9247261} mainly leveraged Generative Adversarial Networks (GANs)~\cite{goodfellow2014generative}
to first warp the reference garment to fit the person's body and then transfer the deformed garment into the source person. 
However, the limited generational capacity of GANs significantly influences the quality of try-on images for these GAN-based methods.

Recently, based on the amazing performance of diffusion models\cite{rombach2022high,10839074,9999553,10444903} in generating high-quality images at high resolutions,
many diffusion-based virtual try-on methods~\cite{Zhu_2023_CVPR_tryondiffusion,chen2023anydoor,morelli2023ladi,baldrati2023multimodal,gou2023taming,kim2024stableviton,choi2024improving} have been proposed to generate natural and realistic try-on images.
For instance, StableVITON~\cite{kim2024stableviton} employed a ControlNet-like~\cite{zhang2023adding} encoder to ensure the fine-grained garment transfer to person data.
IDM-VTON ~\cite{choi2024improving} designed a dual U-Net architecture to respectively encode the person feature and cloth feature, then conducted the cross attention between them to achieve high-quality garment transfer in wild realistic scenarios.
While significant progress has been made, these methods heavily depended on the quality of the inpainting mask to determine the try-on area.
When evaluating more complex try-on data that contains diverse foreground occlusions and person poses, they always fail to restore the non-try-on area.

To alleviate the above issue, TPD~\cite{yang2024texture} and Betterfit~\cite{zhang2024better} proposed mask prediction or correlation modules to dynamically identify precise try-on areas.
However, these methods do not completely eliminate the dependencies of the mask guidance.
Furthermore, BooW-VTON~\cite{zhang2024boow} and AnyDesign~\cite{niu2024anydesign} adopted a mask-free virtual try-on training paradigm.
Though achieving impressive progress, they sometimes struggle to accurately identify try-on areas, 
especially when handling ambiguous or diverse clothing types.

\noindent{\textbf{Video Virtual Try-On.}} 
Recently, inspired by the training paradigm of image virtual try-on, 
several diffusion-based video virtual try-on methods have been proposed~\cite{he2024wildvidfit,fang2024vivid,xu2024tunnel,wang2024gpd,zheng2024vitondit}.
They utilized the inpainting mask to ensure the try-on area and employed the temporal attention module to ensure video coherence.
However, these methods suffer from more severe drawbacks of the mask-based training paradigm due to the increased complexity of video data.
Designing an effective and versatile video virtual try-on paradigm remains both critical and challenging.

In contrast to the aforementioned mask-based paradigm,
we propose a point-enhanced mask-free video virtual try-on method to simultaneously achieve: 1) precise control over the try-on area, 2) accurate preservation of non-try-on regions and 3) temporal coherence of video frames, thus synthesizing more realistic and coherent virtual try-on videos.

\section{The Approach}

\subsection{Preliminary}
\noindent{\textbf{Stable Diffusion.}}
Our PEMF-VTO leverages the Stable Diffusion (SD) ~\cite{rombach2022high},
one of the most widely applied generation models based on the Latent Diffusion Model (LDM).
LDM performs the denoising process in the latent space to conduct a more effective image generation.
Specifically, a VAE encoder $\mathcal{E}(\cdot)$ first converts the image $\mathbf{x}$ into a latent embedding $\mathbf{z}_{0} = \mathcal{E}(\mathbf{x})$.
Then the forward diffusion process is exploited by adding the noise to the latent embedding:
\begin{equation}
    q(\mathbf{z}_t | \mathbf{z}_0) = \mathcal{N}(\mathbf{z}_t ; \sqrt{\bar{\alpha}_t}\mathbf{z}_0, (1-\bar{\alpha}_t)\mathbf{I}),
\end{equation}
where $t \in \{1, ..., T\}$ represents the number of diffusion timesteps and $\{\bar{\alpha}_i\}_{i=1}^{t}$ determines the diffusion schedule.
Finally, the denoising model is trained to predict the added noise of the noisy latent $\mathbf{z}_t$ by the loss constraints of LDM: 
\begin{equation}
    \mathcal{L}_{LDM} = \mathbb{E}_{\mathcal{E}(\mathbf{x}),\mathbf{y},\epsilon\sim\mathcal{N}(0, 1),t}\left[\lVert\epsilon - \epsilon_{\theta}(\mathbf{z}_t, t, \mathbf{y})\rVert_2^2\right],
\label{eq:ldm}
\end{equation}
where $\epsilon_{\theta}$ represents the denoising model and $\mathbf{y}$ is the conditional embedding to control the content of generation. 
\textit{In our task, the feature embeddings derived from the reference garment image should serve as the condition to generate identity-preserved and cloth-changed try-on results.}

\begin{figure*}
	\centering
	\includegraphics[width=0.85\linewidth]{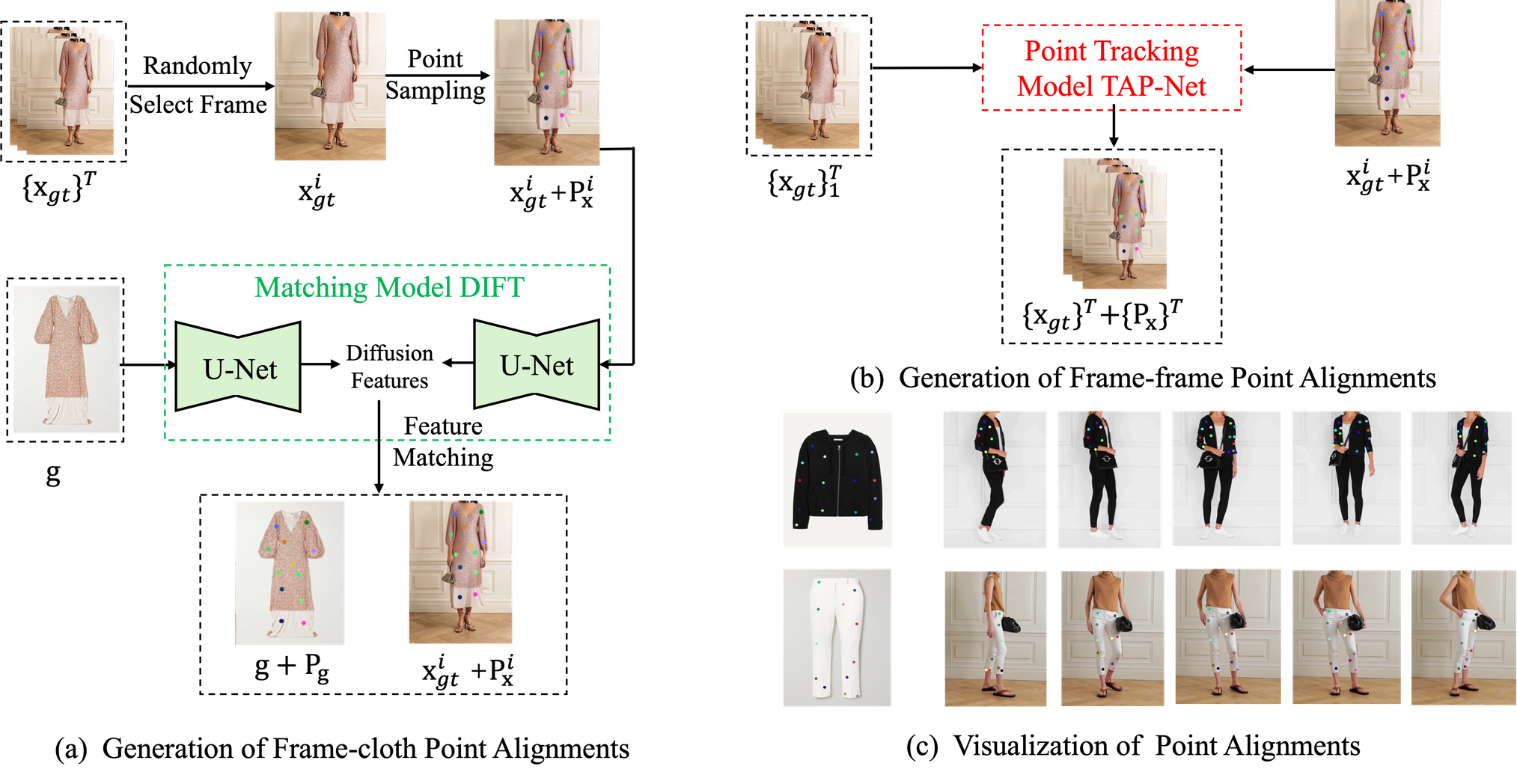}
	\caption{
		The pipeline of the construction for point alignments between video frames and garment images. 
	}\label{fig3}
\end{figure*}

\subsection{Existing Virtual Try-on Paradigms.}
\label{sec:vton_param}
Given a source person video and a reference garment image, our goal is to fluently transfer the garment to a desirable try-on area in the source video, 
thus synthesizing a natural and coherent try-on video.
Current virtual try-on mainly includes two implementation paradigms:
1) Mask-based paradigm ~\cite{fang2024vivid,wang2024gpd,kim2024stableviton,choi2024improving,hu2023animateanyone} and 2) Mask-free paradigm ~\cite{ge2021parser,he2022style,zhang2024boow,niu2024anydesign,issenhuth2020not}.
For the former, it considers the virtual try-on process as a mask-inpainting task, which leverages the pre-acquired agnostic mask to determine the try-on area.
However, when dealing with more complex realistic person videos,
these pre-acquired agnostic masks not only destroy the original spatial information of person actions and postures, 
but also lead to the temporal inconsistency of the try-on area, which has a high risk of generating conflicting and disjointed try-on videos.
For the latter, it first leverages the pre-trained mask-based virtual try-on model to generate the pseudo person video ${\{\textbf{x}_{ps}\}}^{T}$.
Then, the ${\{\textbf{x}_{ps}\}}^{T}$ and the original person video ${\{\textbf{x}_{gt}\}}^{T}$ are paired as training data for the virtual try-on model, enabling the model to achieve accurate try-on results without relying on an agnostic mask. 
This approach prevents the loss of essential spatial-temporal information inherent in the original person data.
However, due to the lack of mask guidance, 
it is more difficult for the mask-free model to identify the accurate try-on area in each frame of the video, 
leading to inconsistencies in the try-on area of the generated videos.

To alleviate issues that brought by above two paradigms, it is worth considering the other explicit guidance to simultaneously achieve 1) \textit{guidance on the garment try-on area}, 2) \textit{preservation of video spatial-temporal information} and 3) \textit{coherence of generated video virtual try-on results}.

\subsection{Point-Enhanced Mask-Free paradigm}
\begin{figure*}[h]
    \centering
    \includegraphics[width=0.95\linewidth]{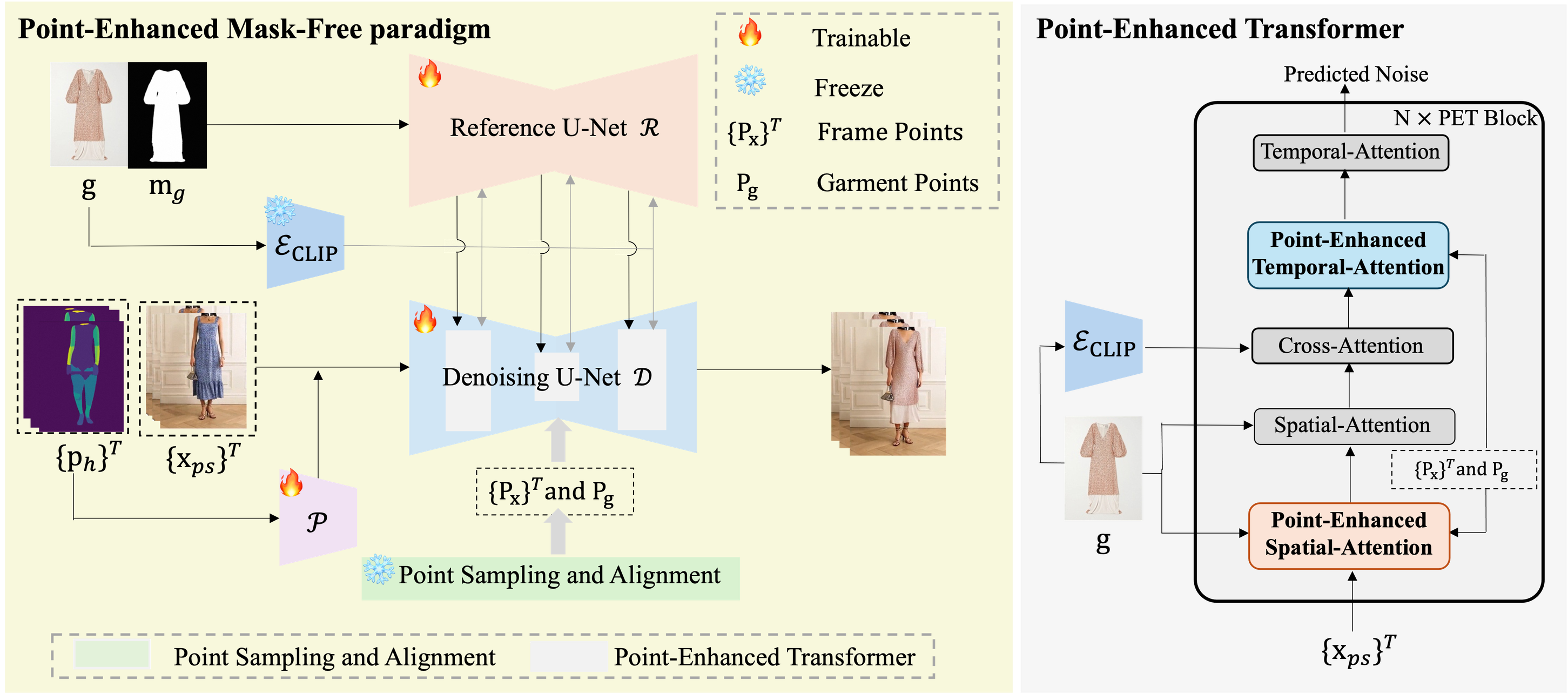}
    \caption{
        The pipeline of our PEMF-VTO framework. 
        It leverages the paired pseudo-person data (${\{\textbf{x}_{ps}\}}^{T}$, ${\{\textbf{x}_{gt}\}}^{T}$) to train a mask-free model, thereby avoiding the loss of spatial-temporal information in the try-on area.
        Besides, based on the pre-acquired alignments between frame points $\{\textbf{P}_{\text{x}}\}^{T}$ and garment points ${\textbf{P}_\textbf{g}}$, 
        a novel point-enhanced transformer is proposed to respectively improve the garment transfer ability and coherence in the try-on area by the point-enhanced spatial attention and point-enhanced temporal attention modules.
    }
    \label{fig2:pipline}
\end{figure*}

\noindent{\textbf{Overview.}} 
In this work, 
we propose a new point-enhanced mask-free paradigm to conduct the video virtual try-on task.
Specifically, we first exploit a pre-trained mask-based try-on model to construct paired pseudo-person training samples,
thus learning a mask-free virtual try-on model.
Then, to further boost the try-on performance,
we leverage the pre-acquired sparse frame-cloth and frame-frame point alignments and integrate a novel Point-Enhanced Transformer (PET) into the mask-free model. 
In PET module, the designed Point-enhanced Spatial Attention (PSA) and Point-enhanced Temporal Attention (PTA)  perform explicit feature alignments of garment image and video frames, as well as alignments between video frames, respectively,
thereby greatly enhancing the garment transfer ability and temporal coherence on more complex realistic human videos. 
The rest of this section will introduce our proposed method in detail.
The pipeline of our PEMF-VTO and PET is shown in Fig.~\ref{fig2:pipline}.
For the pre-trained mask-based model, 
we adopt the open-source video virtual try-on method ViViD ~\cite{fang2024vivid}. 

\subsubsection{Pseudo-Person Data Preparation}
Given a person video ${\{\textbf{x}\}}^{T}$ and a garment image $\textbf{g}$, where $T$ represents the number of video frames, 
we should first obtain the cloth-agnostic video ${\{\textbf{a}\}}^{T}$, agnostic mask sequence ${\{\textbf{m}\}}^{T}$ and human pose ${\{\textbf{p}_{h}\}}^{T}$ of ${\{\textbf{x}\}}^{T}$.
Following ~\cite{fang2024vivid}, the ${\{\textbf{a}\}}^{T}$ and ${\{\textbf{m}\}}^{T}$ are extracted through the human parsing model SCHP~\cite{li2020self,cao2017realtime}.
The pose detection model DensePose~\cite{guler2018densepose} is employed to acquire the pose information ${\{\textbf{p}_{h}\}}^{T}$.
Besides, the garment mask $\textbf{m}_{g}$ of $\textbf{g}$ is segmented by semantic segment model SAM~\cite{kirillov2023segany}.
After acquiring above inputs, 
we leverage the pre-trained mask-based model to perform the virtual try-on operation on the publicly available image and video virtual try-on datasets with randomly selected same-type clothes.
The generated pseudo try-on data ${\{\textbf{x}_{ps}\}}^{T}$ and original person video ${\{\textbf{x}_{gt}\}}^{T}$ are combined as the paired training data.
Then, following ~\cite{niu2024anydesign},
large-scale paired training data is applied to train a powerful mask-free video virtual try-on model.

\subsubsection{Point Sampling and Alignment}
Motivated by recent studies\cite{shi2024dragdiffusion,mou2023dragondiffusion,mou2023diffeditor,chen2024wear}, 
we choose flexible and powerful matching point pairs to enhance both garment transfer accuracy and video coherence.
Therefore, we first acquire the sparse frame-cloth and frame-frame point alignments through the diffusion-based matching model DIFT\cite{tang2023emergent} and point tracking model TAP-Net\cite{doersch2022tap,doersch2024bootstap},
with the points of video frames and garment respectively represented as $\{\textbf{P}_{\text{x}}\}^{T}$ and $\textbf{P}_\textbf{g}$.
The pipeline of our Point Sampling and Alignment is shown in Fig.~\ref{fig3}.
We respectively represent the generation of frame-cloth point alignments and frame-frame point correspondences in Fig.~\ref{fig3} (a) and (b).
Specifically, as shown in Fig.~\ref{fig3} (a),
in the training stage, we first randomly select a frame ${\textbf{x}_{gt}^{i}}$ from the realistic ground truth video ${\{\textbf{x}_{gt}\}}^{T}$.
Then $M$ sparse points $\textbf{P}_\textbf{x}^{i}$ are randomly sampled from the agnostic mask (try-on area) of ${\textbf{x}_{gt}^{i}}$.
In this work, the number of selected points $M \leqslant K$, where $K$ denotes the maximum number of matching correspondences.
After that, DIFT\cite{tang2023emergent} will calculate the semantic-aware matched garment points $\textbf{P}_\textbf{g}$ of $\textbf{P}_\textbf{x}^{i}$ from the garment image $\textbf{g}$.
Besides, as Fig.~\ref{fig3} (b) shown, we exploit the point tracking model TAP-Net\cite{doersch2022tap,doersch2024bootstap} to acquire the frame-frame correspondences based on the select points $\textbf{P}_\textbf{x}^{i}$ in ${\textbf{x}_{gt}^{i}}$.
We also visualize the point alignments in Fig.~\ref{fig3} (c) to confirm the effectiveness os our point sampling and alignment scheme.
In the inference stage, after selecting a frame ${\textbf{x}^{i}}$ from the source video ${\{\textbf{x}\}}^{T}$, users can mark the matching points of ${\textbf{x}^{i}}$ and $\textbf{g}$ 
to achieve a spatially controllable and temporally smooth virtual try-on process.

After acquiring the point alignments, 
we construct two $0-1$ binary masks to respectively represent the positions of frame points $\{\textbf{P}_{\text{x}}\}^{T}$ in ${\{\textbf{x}_{gt}\}}^{T}$ and garment points ${\textbf{P}_\textbf{g}}$ in $\textbf{g}$.
The values at the selected point positions on $\{\textbf{P}_{\text{x}}\}^{T}$ and ${\textbf{P}_\textbf{g}}$ are set to 1.
A max pooling operation $\text{MaxPool}$ is then applied to obtain the masks in lower resolutions, which align with the denoising U-Net $\mathcal{D}$.
Besides, a channel-wise convolution operation is applied to binary masks to expand the receptive field of matching points, 
thus obtaining two soft alignment mask tensors $\{\textbf{M}_{point}^\textbf{x}\}^{T} \in\mathbb{R}^{T\times M\times N_\textbf{x}}$ and $\textbf{M}_{point}^\textbf{g} \in\mathbb{R}^{1\times M\times N_\textbf{g}}$, 
where $N_\textbf{x}$ and $N_\textbf{g}$ are respectively represented the number of pixels in the video frame $\textbf{x}$ and the garment image $\textbf{g}$.

\subsubsection{Point-Enhanced Transformer}

Based on the matched frame points $\{\textbf{P}_{\text{x}}\}^{T}$ and garment points ${\textbf{P}_\textbf{g}}$, as shown in Fig.~\ref{fig2:pipline}, a \textbf{P}oint-\textbf{E}nhanced \textbf{T}ransformer (PET), 
which adds the \textbf{P}oint-enhanced \textbf{S}patial \textbf{A}ttention (PSA) and \textbf{P}oint-enhanced \textbf{T}emporal \textbf{A}ttention (PTA) compared to the transformer layer of denoising U-Net $\mathcal{D}$ in baseline model,
is designed to fully leverage the guidance provided by these point alignments.

\noindent{\textbf{Point-enhanced Spatial Attention.}}
In Fig.~\ref{fig4:pet} (a), we design our point-enhanced spatial attention (PSA) to provide explicit guidance of the try-on area without destroying any spatial information on human movements and postures.
\subsubsection{Point-Enhanced Spatial Attention (PSA)}
Specifically, for a given video frame $\textbf{x}^{i}$, we sample sparse spatial points (e.g., key joints or garment-contact regions) that are \textbf{aligned} with corresponding points in the garment image $\textbf{g}$. These sparse points act as anchors to propagate garment-specific details from $\textbf{g}$ to the full person features in $\textbf{x}^{i}$. To achieve this, we design a \textbf{sparse cross-attention module} (PSA), which computes attention scores between the full person features ($\textbf{Q}^{\text{x}}$) and sparse person point features ($\textbf{K}_{point}^{\text{x}}$), then uses these scores to inject aligned garment features ($\textbf{V}_{point}^{\text{g}}$) into $\textbf{Q}^{\text{x}}$. This explicitly guides the model to focus on the try-on region while preserving spatial coherence in human movements.

The simplified formulation of PSA is:
\begin{equation}
\begin{aligned}
\text{Attention} &= \text{softmax}\left(\text{LN}(\textbf{Q}^{\text{x}}) \cdot \textbf{K}_{point}^{\text{x}}\right) \\
\textbf{Q}^{\text{x}} &= \textbf{Q}^{\text{x}} + \underbrace{\text{Attention} \cdot \textbf{V}_{point}^{\text{g}}}_{\text{Garment Feature Injection}},
\end{aligned}
\label{eq:psa_simple}
\end{equation}
where:

- $\textbf{Q}^{\text{x}} \in \mathbb{R}^{1 \times N_x \times C}$ derived by linear projection of the full person feature $\textbf{F}^{\text{x}}$ (extracted from $\textbf{x}^{i}$).  
    
- $\textbf{K}_{point}^{\text{x}} \in \mathbb{R}^{1 \times M \times C}$ derived by linear projection of sparse person point features $\textbf{F}_{point}^{\text{x}}$. These sparse features are computed via weighted summation of $\textbf{F}^{\text{x}}$ using the soft alignment mask $\textbf{M}_{point}^{\text{x}}$ as weights.  
    
- $\textbf{V}_{point}^{\text{g}} \in \mathbb{R}^{1 \times M \times C}$ derived by linear projection of sparse garment point features $\textbf{F}_{point}^{\text{g}}$. These are similarly computed by weighted summation of garment features (from $\textbf{g}$) using $\textbf{M}_{point}^{\text{g}}$ as weights.

$\text{LN}$ represents layer normalization and $C$ denotes the channel dimension.
The softmax-normalized attention scores quantify how strongly each spatial location in $\textbf{x}^{i}$ should attend to the sparse points and their corresponding garment features. Layer normalization (LN) stabilizes this computation, while the residual update injects garment details into $\textbf{Q}^{\text{x}}$, refining the try-on region.

\begin{figure}
	\centering
	\includegraphics[width=0.9\linewidth]{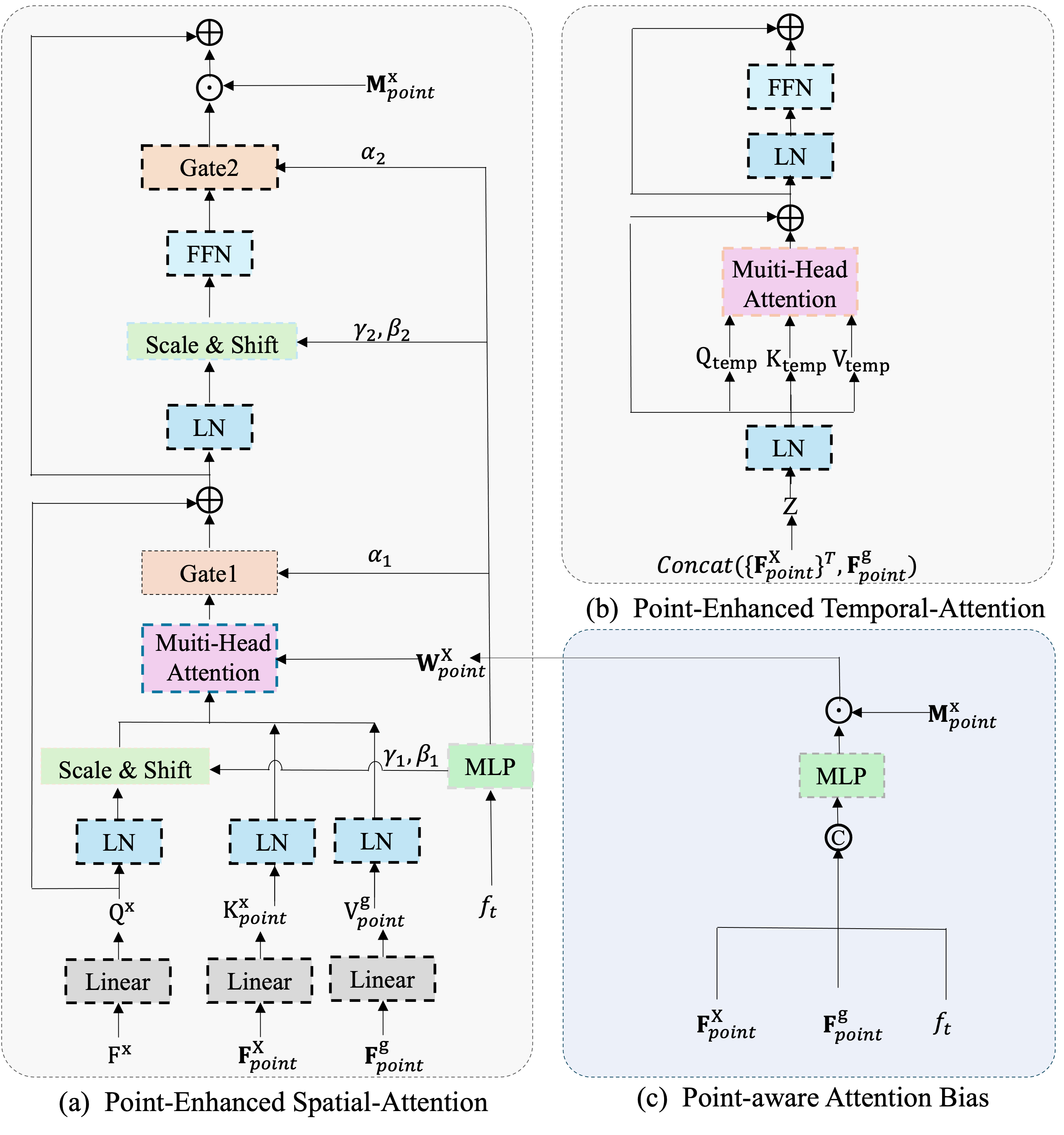}
	\vspace{-0.05in}
	\caption{
		Point-enhanced Spatial Attention (PSA) and Point-enhanced Temporal (PTA) Attention modules. 
	}\label{fig4:pet}
\end{figure}

In Eq.~\ref{eq:psa_simple}, to further facilitate the effective garment feature transfer, 
we design a point-wise attention bias $\textbf{W}_{point}^{\text{x}} \in\mathbb{R}^{1\times N\times M}$ to adaptively adjust the attention scores $\text{Attention}$.
As Fig.~\ref{fig4:pet} (c) shown, $\textbf{W}_{point}^{\text{x}} = \textbf{M}_{point}^\textbf{x} \cdot \text{FFN}(\text{Cat}(\textbf{K}_{point}^{\text{x}}, \textbf{V}_{point}^\textbf{g}, f_{t}))$, 
where $f_{t}$ is the embedding of timestep $t$, and the final attention scores $\text{Attention}^{*} = \text{Attention} + \textbf{W}_{point}^{\text{x}}$. 
Besides, following to the powerful Diffusion Transformer (DiT)~\cite{peebles2023scalable}, 
we leverage the adaptive layer norm block in ~\cite{peebles2023scalable} to increase the learning ability of our PSA  
Specifically, we exploit the  $f_{t}$ to regress the dimension-wise scale parameters $\gamma_{1}$, $\gamma_{2}$ and shift parameters$\beta_{1}$ and $\beta_{2}$ for $\textbf{Q}^{\text{x}}$, 
and the dimension-wise scale parameters $\alpha_{1}$ and $\alpha_{2}$ of residual connections by $\text{MLP}$ layers.
The updated formula of PSA is given as:
\begin{equation}
    \begin{aligned}
        \text{Attention}^{*} &=  \text{softmax}((\gamma_{1}\cdot \text{LN}(\textbf{Q}^{\text{x}}) + \beta_{1}) \cdot \textbf{K}_{point}^{\text{x}} + \textbf{W}_{point}^{\text{x}}) \\
        \textbf{Q}^{\text{x}} &=  \textbf{Q}^{\text{x}} + \alpha_{1} \cdot \text{Attention}^{*} \cdot \textbf{V}_{point}^\textbf{g},
    \end{aligned}
\label{eq:psa}
\end{equation}
Then, to further enhance the controllability of the try-on area, 
the soft alignment mask tensor of person frame $\textbf{M}_{point}^\textbf{x}$ explicitly constrains the updated residual to only work in the surrounding area of point pairs by the following formula:
\begin{equation}
    \begin{aligned}
         \textbf{F}^{\text{x}}_{updated} = \textbf{Q}^{\text{x}} + \textbf{M}_{point}^\textbf{x}\cdot \alpha_{2} \cdot \text{FFN}(\gamma_{2}\cdot \text{LN}(\textbf{Q}^{\text{x}}) + \beta_{2})
    \end{aligned}
\label{eq:psa_mask}
\end{equation}
where the $\text{FFN}$ is the Feed Forward Network.

\noindent{\textbf{Point-Enhanced Temporal Attention (PTA)}}
Traditional temporal attention mechanisms~\cite{hu2023animateanyone,guo2023animatediff} assume pixel-wise correspondence between frames (e.g., same coordinates across frames), which often fails for dynamic human motions with complex deformations or occlusions. To address this, we design a \textbf{Point-Enhanced Temporal Attention (PTA)} module that explicitly leverages the pre-acquired frame-frame point alignments $\{\textbf{M}_{point}^{\text{x}}\}^{T}$ to enforce temporal coherence in the try-on region. This approach ensures feature consistency between corresponding garment parts rather than relying on rigid spatial grids.

The PTA module operates in two stages:

\begin{enumerate}
    \item \textbf{Temporal Feature Construction:}  
    For each video frame $\textbf{x}^i$, we first aggregate its sparse person point features $\textbf{F}_{point}^{\text{x}^i}$ using the soft alignment mask $\textbf{M}_{point}^{\text{x}^i}$. These features are then concatenated across all $T$ frames to form a temporal sequence $\{\textbf{F}_{point}^{\text{x}}\}^{T} \in \mathbb{R}^{T \times M \times C}$. Similarly, we can obtain the garment point features $\textbf{F}_{point}^{\text{g}} \in \mathbb{R}^{1 \times M \times C}$. The combined tensor $\textbf{Z} \in \mathbb{R}^{(T+1) \times M \times C}$ is constructed as:  
    $$
    \textbf{Z} = \text{Concat}(\{\textbf{F}_{point}^{\text{x}}\}^{T},\ \textbf{F}_{point}^{\text{g}})
    $$  
    This design explicitly binds garment details to each frame's motion trajectory.

    \item \textbf{Sparse Temporal Self-Attention:}  
    We apply a sparse self-attention mechanism $\texttt{SelfAttn}$ to $\textbf{Z}$, focusing on interactions between aligned point pairs across frames:  
    $$
    \textbf{Z}_{updated} = 
    \text{Softmax}\left(\textbf{Q}_{\text{temp}} \cdot \textbf{K}_{\text{temp}} \right) \cdot \textbf{V}_{\text{temp}},
    $$  
    where:  
    $\textbf{Q}_{\text{temp}}, \textbf{K}_{\text{temp}}, \textbf{V}_{\text{temp}} \in \mathbb{R}^{(T+1) \times M \times C}$ are linear projections of $\textbf{Z}$.  
    Attention scores are computed only between points that share temporal correspondence guided by $\{\textbf{M}_{point}^{\text{x}}\}^{T}$, ensuring efficient computation and physical plausibility.  
\end{enumerate}

The updated tensor $\textbf{Z}_{updated}$ is split back into temporal person features $\{\textbf{F}_{point}^{\text{x}}\}^{T}_{updated}$, which refine the spatial try-on process in subsequent layers.
This formulation ensures that garment transfer adheres to motion trajectories while suppressing artifacts from misaligned or occluded regions. By integrating explicit point alignments into temporal attention, PTA significantly improves the smoothness of try-on results in dynamic sequences.

\noindent{\textbf{Remark.}} 
The implementation of PEMF-VTO is mainly motivated by the deficiencies of existing virtual try-on paradigms in \ref{sec:vton_param}, 
which can be clearly illustrated by Fig.~\ref{fig1}.
From Fig.~\ref{fig1}, we observe that the mask-based method can not restore the crucial spatial-temporal details from the agnostic mask, 
and the mask-free method struggles to accurately perceive the try-on regions, especially for realistic in-the-wild human videos.
To this end, it is natural and intuitive to leverage more flexible and reliable point alignments ~\cite{chen2024wear,tang2023emergent,doersch2024bootstap} to guide the virtual try-on model, 
thereby obtaining more reasonable and coherent video virtual try-on results.
Specifically, our designed PSA explicitly enhances the precise garment transfer to the try-on area with the cross-attention operation between sparse frame-cloth point pairs.
Besides, the PTA leverages inter-frame matching point pairs to conduct a enhanced temporal attention, thus achieving superior coherence of the try-on video.
The detailed ablation experiments in \ref{sec:abl} will show the effectiveness of our proposed PET module.

\subsection{Training and Inference}

\noindent{\textbf{Training scheme.}} 
The training scheme of PEMF-VTO comprises three stages:
\textbf{(1) Stage 1}: the denoising U-Net $\mathcal{D}$ is initialized as a 2D inpainting model, and only a single frame of pseudo-person data is taken as training data,
which enables the model to perceive and transfer the reference garment to a reasonable try-on area. 
The parameters of pose encoder $\mathcal{P}$, reference U-Net $\mathcal{R}$ and denoising U-Net $\mathcal{D}$ are updated in this stage.  
\textbf{(2) Stage 2}: we initialize the temporal attention module with the parameters in ~\cite{guo2023animatediff} and exploit both image and video data to only finetune 
it, thus enhancing the temporal coherence of generated try-on results. 
\textbf{(3) Stage 3}: since most of the training samples have been well learned through the first and second stages,
we first leverage the mask-free model of the second stage to construct hard-paired pseudo-person training samples with lower generation performance (\textit{i.e.}~SSIM\textless 0.75).
Then, to ensure the PSA and PTA can really promote the temporal consistent and controllable garment transfer,
we only train the PSA module and the PTA module with these hard training samples.
The learning objectives of the three training stages are the same LDM loss in Eq.~\ref{eq:ldm}.

\noindent{\textbf{Inference scheme.}}
Compared to existing video virtual try-on methods, our PEMF-VTO provides a more intelligent and user-friendly virtual try-on experience.
Specifically, according to most human videos with simple action postures, 
it is enough to achieve accurate and desirable try-on results even without the guidance of point alignments.
However, when dealing with more complex human videos (\textit{e.g.} street dance videos) or aiming at more controllable try-on results,
users can manually click on the matching point pairs between a randomly selected single video frame and reference garment image,
thereby acquiring a more coherent and natural generation.
Besides, to enable the smooth and consistent try-on results for long videos, 
we follow ~\cite{fang2024vivid} to employ a sliding window strategy in the inference stage. 


\section{Experiments}

\begin{table*}
    \centering
    \renewcommand\arraystretch{1.0}
    \resizebox{0.95\linewidth}{!}{
    \begin{tabular}{l|cc|cc|cc|ccc|ccc }
    \toprule

     \multirow{3}{*}{\textbf{Method}}  & 
     \multicolumn{6}{c|}{Unpaired}  & \multicolumn{6}{c}{Paired} \\
     \cmidrule{2-7}  \cmidrule{8-13}
    & \multicolumn{2}{c|}{\textbf{VVT}}  & \multicolumn{2}{c|}{\textbf{ViViD}}  & \multicolumn{2}{c|}{\textbf{TikTok}} & \multicolumn{3}{c|}{\textbf{VVT}}  & \multicolumn{3}{c}{\textbf{ViViD}} \\
     \cmidrule{2-4}  \cmidrule{5-7} \cmidrule{8-10} \cmidrule{11-13}
     
      & \textbf{VFID$_{I}$}\(\downarrow\) & \textbf{VFID$_{R}$}\(\downarrow\) & \textbf{VFID$_{I}$}\(\downarrow\) & \textbf{VFID$_{R}$}\(\downarrow\) & \textbf{VFID$_{I}$}\(\downarrow\) & \textbf{VFID$_{R}$}\(\downarrow\) & \textbf{SSIM}\(\uparrow\) & \textbf{LPIPS}\(\downarrow\) & \textbf{VFID$_{I}$}\(\downarrow\)  & \textbf{SSIM}\(\uparrow\) & \textbf{LPIPS}\(\downarrow\) & \textbf{VFID$_{I}$}\(\downarrow\)  \\
    \midrule
    PF-AFN${}^{*}$~\cite{ge2021parser}  & 5.12  & 0.125 & 43.22  & 1.87 & 43.38  & 7.16 & 0.856 & 0.233 &  7.33 &  0.776 & 0.183  &  39.28\\
    Flow-Style${}^{*}$~\cite{he2022style} & 4.79  & 0.097 & 41.93  & 1.69 & 42.28 & 11.38 & 0.854 & 0.241 &  7.64 & 0.781  & 0.169  &  38.64\\  
    StableVITON${}^{*}$~\cite{kim2024stableviton}  & -  & - & 36.90 & 0.91 & - & -  & 0.902 & 0.078 & 3.54  & 0.802 & 0.134 & 34.24 \\ 
    IDM-VTON${}^{*}$~\cite{choi2024improving} & 2.77  & 0.027 & 25.50  & 0.72 & 38.62 & 5.97  & 0.896 & 0.079 & 3.91  & 0.823 & 0.116 & 20.08 \\ 
    CatVTON${}^{*}$~\cite{chong2024catvton}  & 2.49  & 0.016 & 22.65  & 1.14 & 39.67 & 6.53  & 0.899 & 0.082 & 3.75   &  0.834 & 0.089 & 17.73 \\ \hline
    WildVidFit${}^{\dag}$~\cite{he2024wildvidfit}   & 4.20  & - & -  & - & - & -  & -  & - & -  & - & - & - \\
	GPD-VVTO${}^{\dag}$~\cite{wang2024gpd}  & 1.28  & - & -  & - & - & -  & \textbf{0.928}  &0.056 & -  & - & - & - \\ 
	ViViD${}^{\dag}$~\cite{fang2024vivid} & 3.99  & 0.041 & 21.80  & 0.82 & 46.73 & 6.94  & 0.822  &0.107 & 3.78  & 0.803 & 0.122 & 17.29 \\ 
    CatV${}^{2}$TON${}^{\dag}$ ~\cite{chong2025catv2ton} & 1.90  & 0.014 & 19.51  & 0.53 & 43.62 & 6.18  & 0.900  &0.039 & 1.78  & 0.873 & 0.064 & 13.60 \\ \hline
	\cellcolor{LightCyan}\textbf{PEMF-VTO} & \cellcolor{LightCyan}\textbf{0.95} & \cellcolor{LightCyan}\textbf{0.007} & \cellcolor{LightCyan}\textbf{16.67}  &  \cellcolor{LightCyan}\textbf{0.34} & \cellcolor{LightCyan}\textbf{31.62} & \cellcolor{LightCyan}\textbf{2.05} & \cellcolor{LightCyan}0.915 & \cellcolor{LightCyan}\textbf{0.035} & \cellcolor{LightCyan}\textbf{0.87}  &  \cellcolor{LightCyan}\textbf{0.911} & \cellcolor{LightCyan}\textbf{0.040} & \cellcolor{LightCyan}\textbf{7.54} \\ 
    \bottomrule
    \end{tabular}
}
\caption{Quantitative results on the VVT, ViViD and TikTok datasets. The best are marked in bold.
The $*$ and $\dag$ respectively denote image and video virtual try-on methods.}  
\label{tab:video_unpair_agn}
\end{table*}

\begin{table}[]
    \centering
    \renewcommand\arraystretch{1.2}
    \resizebox{0.99\linewidth}{!}{
    \begin{tabular}{l|ccc|ccc}
    \toprule

     \multirow{2}{*}{\textbf{Method}}  & 
     \multicolumn{3}{c|}{\textbf{VITON-HD}}  & \multicolumn{3}{c}{\textbf{DressCode}} \\
     \cmidrule{2-4}  \cmidrule{5-7}
     
      & \textbf{SSIM}\(\uparrow\) & \textbf{LPIPS}\(\downarrow\) & \textbf{FID}\(\downarrow\)  & \textbf{SSIM}\(\uparrow\) & \textbf{LPIPS}\(\downarrow\) & \textbf{FID}\(\downarrow\)  \\
    \midrule
    PF-AFN${}^{*}$~\cite{ge2021parser}  & 0.857 & 0.142 &  17.28 & 0.878 & 0.102 & 20.51  \\ 
    Flow-Style${}^{*}$~\cite{he2022style}  & 0.860 & 0.133 &  16.84 & 0.882 & 0.094 & 19.24  \\ 
	LaDI-VTON${}^{*}$~\cite{morelli2023ladi}   & 0.871 & 0.094 & 13.01   &  0.915 & 0.062 & 16.71  \\  
    StableVITON${}^{*}$~\cite{kim2024stableviton}  & 0.888 & 0.073 & -  & - & - & -  \\ 
    IDM-VTON${}^{*}$~\cite{choi2024improving}  & 0.870 & 0.102 & \textbf{6.29}& 0.920 & 0.062 & 8.64 \\ \hline
    WildVidFit${}^{\dag}$~\cite{he2024wildvidfit}   & 0.883  & 0.077 & 8.67  & \textbf{0.928} & 0.043 & 12.48   \\
	GPD-VVTO${}^{\dag}$~\cite{wang2024gpd}   & 0.891  &0.070 & 8.57  & 0.924 & 0.045 & 4.18  \\ 
    ViViD${}^{\dag}$~\cite{fang2024vivid}   & 0.881  &0.089 & 8.67 & 0.907 & 0.070 & 8.28 \\ \hline
	\cellcolor{LightCyan}\textbf{PEMF-VTO}  & \cellcolor{LightCyan}\textbf{0.894} & \cellcolor{LightCyan}\textbf{0.062} & \cellcolor{LightCyan}6.86  & \cellcolor{LightCyan}0.927 & \cellcolor{LightCyan}\textbf{0.034} & \cellcolor{LightCyan}\textbf{3.41}  \\ 
    \bottomrule
    \end{tabular}
}
\caption{Quantitative results on the VITON-HD and DressCode datasets. The best are marked in bold.
The $*$ and $\dag$ respectively denote image and video virtual try-on methods.}
\label{tab:image_agn}
\end{table}

\subsection{Dataset and Experimental Setting}
\noindent\textbf{Datasets.}
Following our baseline backbone ~\cite{fang2024vivid},
we train our model on two image datasets, VITON-HD~\cite{choi2021viton} and DressCode~\cite{morelli2022dresscode}, and one video dataset, ViViD~\cite{fang2024vivid}.
For the video virtual try-on, following CatV${}^{2}$TON~\cite{chong2025catv2ton},
the VVT~\cite{dong2019fw} dataset and a testset of ViViD~\cite{fang2024vivid} are selected as the evaluation datasets.
Besides, referring to ~\cite{he2024wildvidfit}, a more realistic and challenging TikTok ~\cite{jafarian2021learning} dataset is chosen to further illustrate the superiority and generalization of our method.
For the image virtual try-on, 
comparative experiments are conducted on the test sets of VITON-HD, DressCode and StreetVTON~\cite{cui2023street-tryon} datasets.
All datasets are open-sourced for wide research purposes.
\begin{itemize}
    \item VITON-HD includes 13,679 pairs of upper-body model and garment images, with 2,032 pairs choosing as testing data. 
    \item DressCode comprises 15,363 pairs of full-body models corresponding to upper-body garments, 8,951 for lower-body garments, and 2,947 for dresses.
    \item StreetVTON is a subset collected from DeepFashion to conduct wild image try-on task, containing 2,089 person images with complex backgrounds and body postures.
    \item VVT consists of 791 video clips with a resolution of $256 \times 192$, 
    which is split into a training set of 661 clips and a test set of 130 clips.
    \item ViViD dataset includes 9,700 model videos and garment images with a resolution of $832 \times 624$. 
    Following ~\cite{fang2024vivid}, 
    it is divided into 7,759 videos for the training set and 1,941 videos for the test set. 
    Given the impracticality of testing on thousands of videos in terms of time and cost, inspired by ~\cite{chong2025catv2ton},
    we selected the same 180 videos with CatV${}^{2}$TON~\cite{chong2025catv2ton} from the ViViD test set (60 dresses, 60 uppers, and 60 bottoms), 
    each containing 64 consecutive frontal frames, to form a effective and reasonable testset.
    \item TikTok comprises 340 dance videos that captures a single person dancing with complex body movements. 
    We randomly selected 45 videos from the TikTok dataset as the test set to further illustrate the generalization and robustness of our method.
    The garment images of this test set are randomly selected from the uppers in ViViD dataset.
    The virtual try-on for this dataset is more challenging than simple human model videos in VVT and ViViD. 
\end{itemize}

\noindent\textbf{Metrics.} 
Following previous methods~\cite{fang2024vivid,wang2024gpd,chong2025catv2ton},
both frame-level metrics LPIPS~\cite{zhang2018unreasonable}, SSIM~\cite{wang2004image}, FID~\cite{heusel2017gans} and video-level metric VFID~\cite{unterthiner2018towards} metric with I3D~\cite{carreira2017quo} and ResNext are adopted for comprehensive evaluation.

\begin{figure*}
	\centering
	\includegraphics[width=0.95\linewidth]{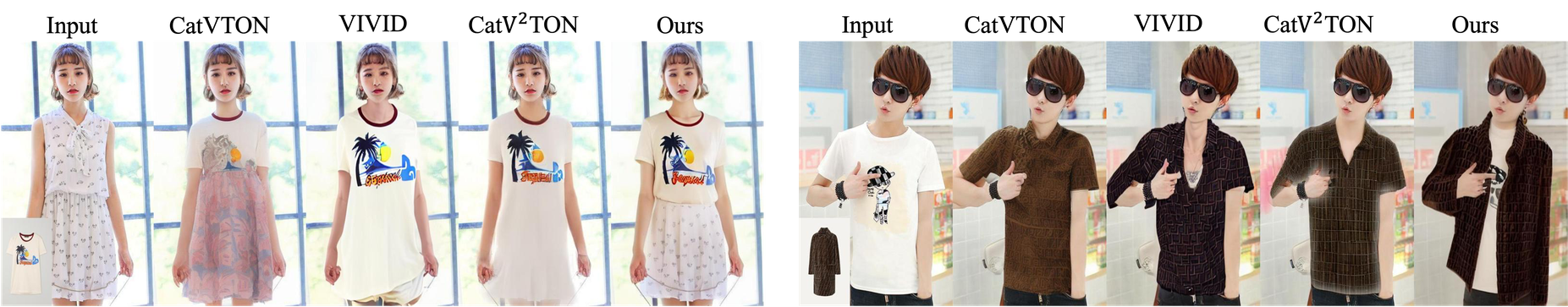}
	\vspace{-0.05in}
	\caption{
    Qualitative comparison on StreetVTON dataset. 
	}\label{fig5}
\end{figure*}

\begin{figure*}
	\centering
	\includegraphics[width=0.95\linewidth]{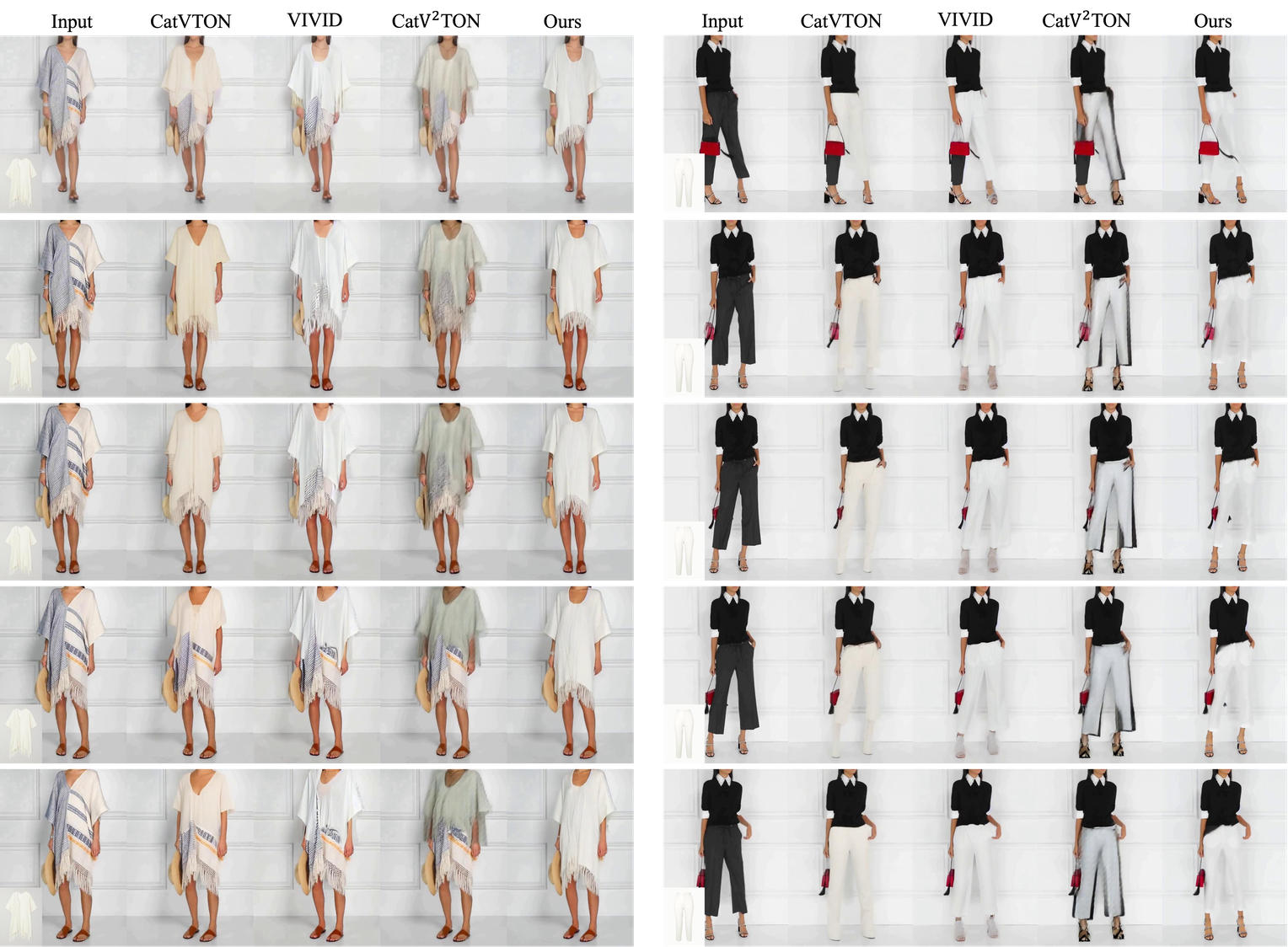}
	\vspace{-0.05in}
	\caption{
    Qualitative comparison on ViViD dataset. 
	}\label{fig6}
\end{figure*}

\begin{figure*}
	\centering
	\includegraphics[width=0.9\linewidth]{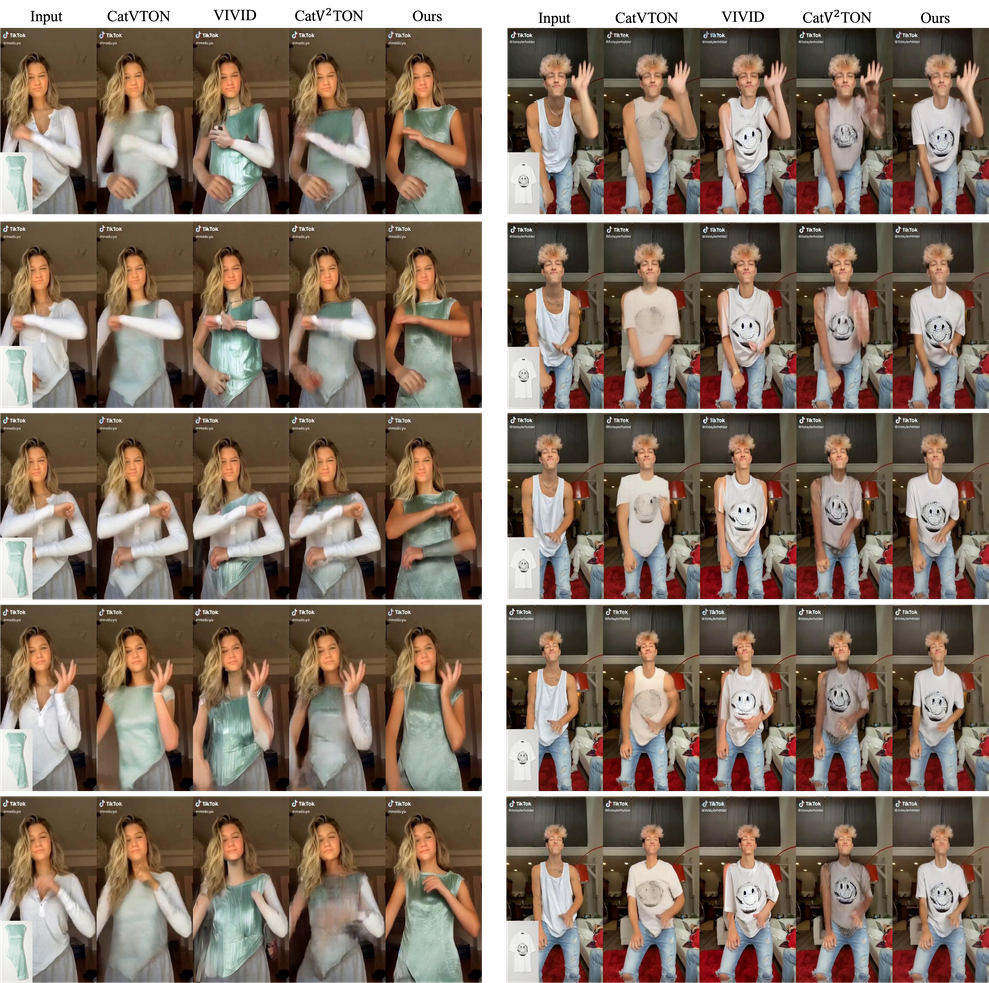}
	\vspace{-0.05in}
	\caption{
    Qualitative comparison on TikTok dataset. 
	}\label{fig7} 
\end{figure*}

\noindent\textbf{Implementation Details.}
We initialize the denoising U-Net $\mathcal{D}$ and reference U-Net $\mathcal{R}$ with the weights from Stable Diffusion-1.5.
The temporal module is initialized with the weights from the motion module of ~\cite{guo2023animatediff}, and the CLIP image encoder is the same as the baseline model~\cite{fang2024vivid}.
During training, all data is resized to a uniform resolution of $512 \times 512$.
Experiments are conducted on 8 Nvidia H100 GPUs with a learning rate of 5e-5.
The numbers of training iterations in the three stages are 40000, 40000 and 20000 respectively.
Since images can be considered as a single frame of videos, both image and video data are utilized during the three training stages, with the proportions of video data being 0.3, 0.9 and 0.5 for each stage respectively. 
In each iteration, only one form of training data is chosen, either video or image. 
With the selection of image datasets, training is conducted with a batch size of 128. 
In contrast, the selection of video datasets involves training the model with 16-frame video sequences and a batch size of 8.
To enable a fair comparison with mask-based methods, the agnostic mask can be treated as a specific garment to participate in the training of our PEMF-VTO, 
accompanied by 20\% ratios.
The maximum number of points $K$ is set to 16.

\subsection{Quantitative Results}

\noindent\textbf{Video Virtual Try-on.}
To achieve a fair and comprehensive comparison with previous methods, 
we evaluate our method on two test settings for the video virtual try-on task.
The unpaired setting means that the virtual try-on model should substitute the garment of the input source person video with a different garment.
The paired setting is defined as reconstructing the person video by providing an agnostic person video and its original garment.
In Tab. \ref{tab:video_unpair_agn}, four video try-on methods WildVidFit~\cite{he2024wildvidfit}, GPD-VVTO~\cite{wang2024gpd}, ViViD ~\cite{fang2024vivid} and CatV${}^{2}$TON ~\cite{chong2025catv2ton} are chosen as the video-based comparative methods.
Besides, many relatively high-performing image-based virtual try-on methods are also exploited to enrich the experimental comparisons.
As Tab. \ref{tab:video_unpair_agn} shown, our PEMF-VTO significantly outperforms all image-based and video-based methods in all metrics of two different settings.
The superior performance on two simple model video datasets VVT and ViViD and one realistic in-the-wild dataset TikTok further demonstrates the generalization ability of our method.

\noindent\textbf{Image Virtual Try-on.}
To further illustrate the effectiveness and robustness of our proposed method, for the image virtual try-on task,
we compare our method with both image-based and video-based virtual try-on methods in the paired setting.
As shown in Tab. \ref{tab:image_agn},
although our approach is mainly designed for video virtual try-on, 
it outperforms two video-based try-on methods across all metrics and achieves comparable performance with SOTA image-based virtual try-on methods, 
This is primarily due to the lower quality of video frames used during the training stage of our PEMF-VTO, which negatively impacts the model's image generation capability.
These experimental results show that
our method can also successfully generalize to the image virtual try-on task to show competitive performance.

\subsection{Qualitative Results}

As shown in Fig.~\ref{fig5}, Fig.~\ref{fig6} and Fig.~\ref{fig7}, we show the visualization comparisons of our method 
with the SOTA image-based method CatVTON~\cite{chong2024catvton} and 
two video-based method CatV${}^{2}$TON~\cite{chong2025catv2ton} and ViViD~\cite{fang2024vivid}.
Specifically, 
our generated results achieve significant visual fidelity to the reference garment in the spatial dimension and content consistency in the temporal dimension in all datasets.
It is worth noting that our model is only trained on simple model images and videos and obtains superior try-on results on more challenging StreetVTON and TikTok datasets, 
which demonstrates the effectiveness and generalization of our method.

\begin{figure}[t]
	\centering
	\includegraphics[width=0.9\linewidth]{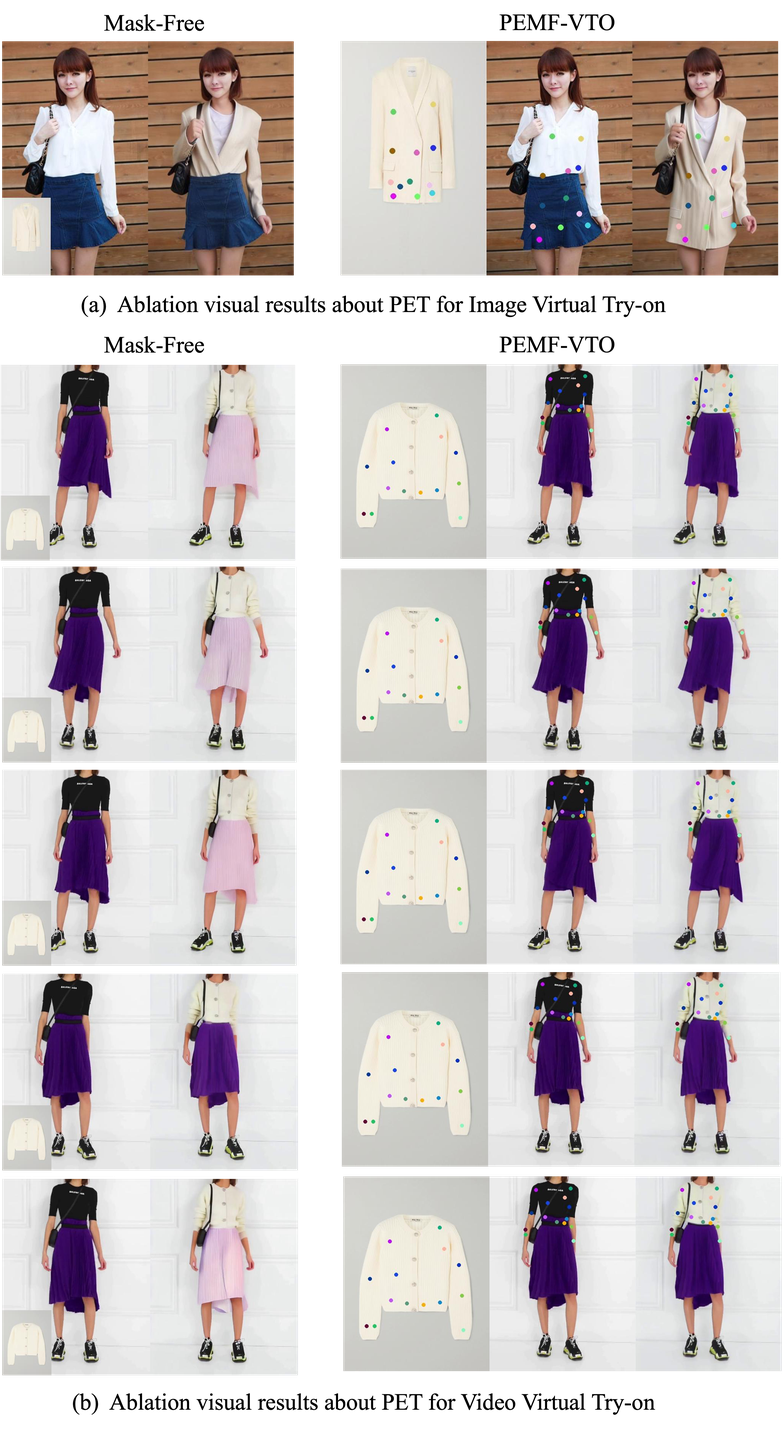}
	\caption{
    Qualitative results for the Ablation visual results about PET. The visualizations of virtual try-on are respectively from StreetVTON and ViViD datasets.
	}\label{fig8}
\vspace{-0.1in}
\end{figure}

\begin{table}[t]
\centering
\footnotesize
\renewcommand\arraystretch{1.1}
\resizebox{0.99\linewidth}{!}{%
\begin{tabular}{l|cc|cc|cc}
    \toprule[1.25pt]
    \multirow{3}{*}{\textbf{Variants}} & \multicolumn{4}{c|}{\textbf{Realistic}} & \multicolumn{2}{c}{\textbf{Pseudo}}   \\
    \cmidrule(rl){2-7}
    &  \multicolumn{2}{c|}{\textbf{ViViD}} & \multicolumn{2}{c|}{\textbf{TikTok}} &  \multicolumn{2}{c}{\textbf{ViViD}}      \\ 
    \cmidrule(rl){2-3}\cmidrule(rl){4-5} \cmidrule(rl){6-7}  
    & \textbf{VFID$_{I}$}\(\downarrow\) & \textbf{VFID$_{R}$}\(\downarrow\)     & \textbf{VFID$_{I}$}\(\downarrow\) & \textbf{VFID$_{R}$}\(\downarrow\)    & \textbf{SSIM}\(\uparrow\)     & \textbf{LPIPS}\(\downarrow\)   \\ \hline
ViViD               &   21.80&        0.82&         46.73&          6.94& 0.840    & 0.114                 \\
Mask-Free               &   19.32&        0.57&        38.96 &        4.38  & 0.857    & 0.083               \\   
PEMF-VTO                 &   \textbf{16.67}&        \textbf{0.34}&        \textbf{31.62} &       \textbf{2.05}   & \textbf{0.867}    & \textbf{0.073}             \\\midrule[1.25pt]
base + PSA          &   \textbf{17.86}&        \textbf{0.42}&         \textbf{33.69}&       \textbf{2.68}& \textbf{0.863}          & \textbf{0.075}            \\    
\textit{w/o} $\textbf{M}_{point}^\textbf{x}$             &   18.44&        0.47&         34.72&           3.14& 0.860          &  0.078                \\
\textit{w/o} $\textbf{W}_{point}^{\text{x}}$              &   18.23&        0.45&         34.17&           2.91& 0.862          &  0.076            \\  \midrule[1.25pt] 
base + PTA              &   \textbf{18.03}&        \textbf{0.44}&         \textbf{34.78}&        \textbf{2.91}& \textbf{0.864}          &   \textbf{0.076}         \\ 
\textit{w/o} cat $\textbf{F}_{point}^\textbf{g}$           &   18.52&        0.48&         35.61&           3.17& 0.861          &  0.079           \\ \midrule[1.25pt]
\end{tabular}%
}
\caption{Ablation studies for our PSA and PTA modules on realistic ViViD and TikTok datasets and pseudo ViViD dataset.}
\label{tab:abl_pet}
\vspace{-0.25in}
\end{table}

\begin{figure}[t]
	\centering
	\includegraphics[width=0.89\linewidth]{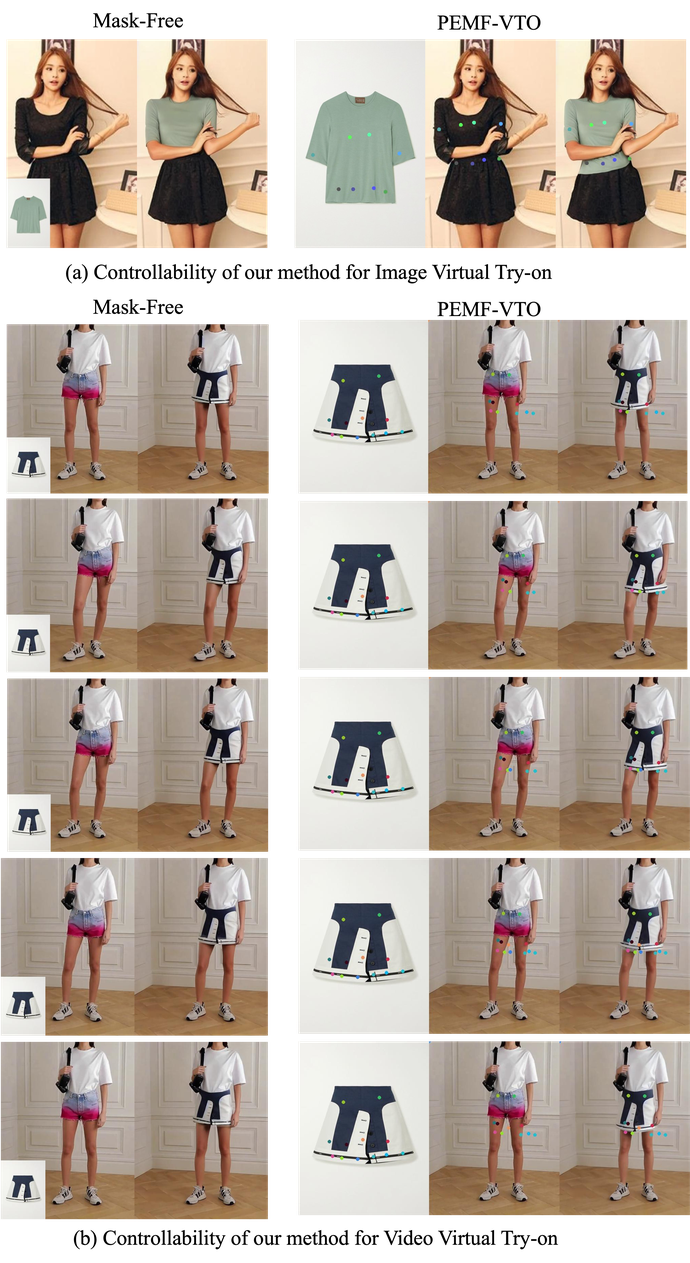}
	\caption{
    Qualitative results for the Controllability of our method. The visualizations of virtual try-on are respectively from StreetVTON and ViViD datasets.
	}\label{fig9}
\vspace{0.25in}
\end{figure}

Besides, in Fig.~\ref{fig1} and Fig.~\ref{fig8}, 
it is clear that existing mask-based and mask-free virtual try-on paradigms have obvious deficiencies, especially for realistic video data with complex scene changes and body movements.
Concretely, the mask-based method can not recover the important spatial-temporal information due to the inaccurate agnostic mask.
The mask-free method may get confused about how to determine a reasonable and coherent try-on area.
Differently, our PEMF-VTO can simultaneously achieve the precise determination of try-on area and the video coherence without the guidance of the agnostic mask.
Furthermore, as Fig.~\ref{fig9} shown, 
our method has the potential to conduct a more flexible and controllable try-on process through the explicit guidance of the point alignments.

\subsection{Ablation Studies}
\label{sec:abl}
In this section, we design different variants to perform a detailed analysis of our PET module. 
To increase the credibility,
in Tab. \ref{tab:abl_pet}, 
we also synthesized 430 pseudo video pairs with the points alignments from the ViViD testing dataset.
Thus, the LPIPS and SSIM metrics can be applied to further evaluate the generation quality.

\noindent\textbf{Point-Enhanced Transformer.}
As shown in Tab. \ref{tab:abl_pet}, compared to the baseline (Mask-Free variant),
our PEMF-VTO obtains obvious performance improvement to acquire more natural and coherent try-on results.

\noindent\textbf{Point-Enhanced Spatial Attention.}
Based on the frame-cloth point alignments,
the PSA is designed to provide explicit guidance to transfer the garment into the desirable try-on area, 
which significantly improves the performance of our baseline model as Tab. \ref{tab:abl_pet} shown.
Besides, in the upper part of Tab. \ref{tab:abl_pet}, 
we analyze the effectiveness of soft alignment mask $\textbf{M}_{point}^\textbf{x}$ and point-wise attention bias $\textbf{W}_{point}^{\text{x}}$.
Our PSA obtains performance decrease when removing the two designs, which illustrates the reasonability of PSA.

\noindent\textbf{Point-Enhanced Temporal Attention.}
The complex human actions at the temporal level always lead to the discontinuity of the try-on video.
To this end, we design the PTA to enhance the coherence of the try-on area between different frames.
As Tab. \ref{tab:abl_pet} shown, the PTA brings clear improvements to the baseline model.
Besides, the performance of PTA decreases when only selecting the sparse person point features $\textbf{F}_{point}^\textbf{x}$ 
(\textit{i.e.} \textit{w/o} $\textbf{F}_{point}^\textbf{g}$) as the input of temporal-level sparse self-attention module $\texttt{SelfAttn}$ in PTA, 
which further verifies the effectiveness of our PTA.

\begin{figure}[t]
	\centering
	\includegraphics[width=0.85\linewidth]{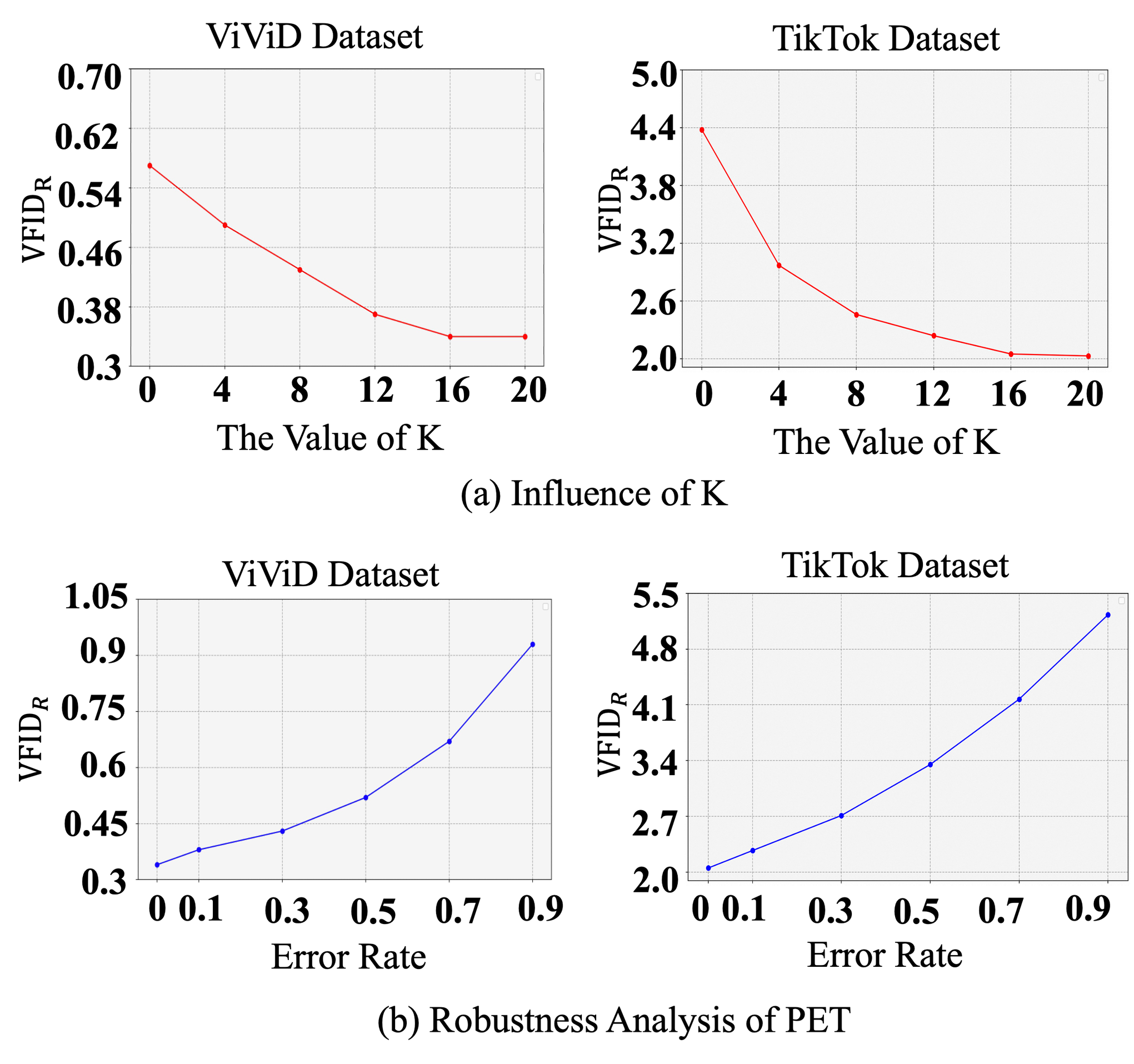}
	\caption{
		(a) Ablation studies for different $K$ and (b) Robustness Analysis for different error rates of point alignments. 
        We conduct the experiments on the ViViD and TikTok datasets. 
	}\label{fig10}
\vspace{-0.25in}
\end{figure}

\noindent\textbf{Different maximum number of points.} 
In Fig.~\ref{fig10} (a), we investigate the effects of the maximum number of points $K$ on the performance of the ViViD and TikTok datasets.
As $K$ increases, the FID and VFID metrics first improve and then stabilize. 
It can be attributed to the repetition of more sampling points.
Therefore, we choose $K=16$ in this paper.

\noindent\textbf{Robustness Analysis of PET.} 
The pre-acquired point alignments have the probability of error which may lead to a negative impact on the try-on performance.
Therefore, to verify the robustness of our PET,
we first randomly perturb a specific proportion of the pre-acquired point alignments to obtain the point pair guidance with different error rates.
In Fig.~\ref{fig10} (b), if the error rate is low (\textit{i.e.} \textless 20\%), our method shows impressive robustness to achieve similar improvement with all correct variants.
Besides, even though the error rate is nearly 70\%, the VFID$_{R}$ of our PEMF-VTO on the TikTok dataset also outperforms the base mask-free model, 
thereby significantly reflecting the reasonability and robustness of our proposed PET module.

\section{Limitation}

Although our proposed PEMF-VTO has strongly enhanced
the accurate garment transfer ability in both spatial
and temporal dimensions, our method may generate incorrect
try-on results when dealing with longer input video
with more ambiguous garment types. Besides, the ability to
control and edit the try-on area should be further improved.
In the future, we will construct more diverse person-pseudo
training data to continually train our model, thereby promoting
the continuous upgrading and evolution of our video
virtual try-on model to achieve better performance on more
challenging and realistic video data.

\section{Conclusion}
To alleviate the deficiencies of existing virtual try-on paradigms and synthesize more realistic and coherent video try-on results,
in this work, we propose a novel Point-Enhanced Mask-Free Video Virtual Try-On method (PEMF-VTO).
Concretely, we first leverage the pre-trained try-on model to construct paired pseudo-person training samples to learn a mask-free try-on model.
Then, based on the pre-acquired sparse alignments, 
the Point-enhance Spatial Attention (PSA) and Point-enhance Temporal Attention (PTA) are designed 
to improve the garment transfer ability and coherence of more complex realistic human video. 
Our PEMF-VTO can simultaneously
achieve
1) the accurate transfer of reference garment 2) the preservation of non-try-on areas and 3) the continuity of generated video frames.
Extensive quantitative and qualitative experimental results clearly show the effectiveness of our method.

\bibliographystyle{splncs04}
\bibliography{main}

\end{document}